\documentclass[12pt]{article}
\usepackage{amsfonts}
\usepackage{a4wide}
\usepackage{amsmath,amscd,amsthm,a4,amssymb}

\begin{document}

\begin{center}
{\large \textbf{MEASURE THEORETIC RESULTS FOR APPROXIMATION BY NEURAL
NETWORKS WITH LIMITED WEIGHTS}}

\

\textbf{\ Vugar E. Ismailov}\footnote{%
Corresponding author at: The Institute of Mathematics and Mechanics, 
9 B. Vahabzadeh str., AZ1141, Baku, Azerbaijan; E-mail:
vugaris@mail.ru}$^{1}$\textbf{\ and Ekrem Savas}$^{2}$ \vspace{4mm}

$^{1}${Institute of Mathematics and Mechanics, Baku,
Azerbaijan} \vspace{1mm}

$^{2}${Department of Mathematics, Istanbul Ticaret University, Istanbul,
Turkey} \vspace{1mm}
\end{center}

\textbf{Abstract.} In this paper, we study approximation properties of
single hidden layer neural networks with weights varying on finitely many
directions and thresholds from an open interval. We obtain a necessary and
at the same time sufficient measure theoretic condition for density of such
networks in the space of continuous functions. Further, we prove a density
result for neural networks with a specifically constructed activation
function and a fixed number of neurons.

\bigskip



\bigskip

\bigskip

\begin{center}
{\large \textbf{1. Introduction}}
\end{center}

For the past 30 years, the topic of artificial neural networks has been a
vibrant area of research. Nowadays, neural networks are being successfully
applied in areas as diverse as computer science, finance, medicine,
engineering, physics, etc. Perhaps the greatest advantage of neural networks
is their ability to be used as an arbitrary function approximation
mechanism. In this paper, we are interested in questions of density (or
approximation with arbitrary accuracy) of a single hidden layer perceptron
model in neural networks. A single hidden layer perceptron model with $r$
units in the hidden layer and input $\mathbf{x}=(x_{1},...,x_{d})$ evaluates
a function of the form

\begin{equation*}
\sum_{i=1}^{r}c_{i}\sigma (\mathbf{w}^{i}\mathbf{\cdot x}-\theta _{i}),\eqno%
(1.1)
\end{equation*}%
where the weights $\mathbf{w}^{i}$ are vectors in $\mathbb{R}^{d}$, the
thresholds $\theta _{i}$ and the coefficients $c_{i}$ are real numbers and \
the activation function $\sigma $ is a univariate function, which is
considered to be continuous in the present paper. For various activation
functions $\sigma $, it has been proved in a number of papers that one can
approximate well to a given continuous function from the set of functions of
the form (1.1) ($r$ is not fixed!) over any compact subset of $\mathbb{R}%
^{d} $, $d\geq 1$. In other words, the set

\begin{equation*}
\mathcal{M}(\sigma )=span\text{\ }\{\sigma (\mathbf{w\cdot x}-\theta ):\
\theta \in \mathbb{R}\text{, }\mathbf{w}\in \mathbb{R}^{d}\}
\end{equation*}%
is dense in the space $C(\mathbb{R}^{d})$ in the topology of uniform
convergence on all compacta (see, e.g., \cite{2,3,Cost,4,7,10,25}). More
general result of this type belongs to Leshno, Lin, Pinkus and Schocken \cite%
{13}. They proved that the necessary and sufficient condition for any
continuous activation function to have the density property is that it not
be a polynomial. This result shows the efficacy of the single hidden layer
perceptron model within all possible choices of the activation function $%
\sigma $, provided that $\sigma $ is continuous.

We recall that the property of a network to approximate any multivariate
function on any compact subset of $d-$dimensional space is called the
universal approximation property. As mentioned above single hidden layer
networks with a nonpolynomial activation function possess this property.
There are several results (see, e.g., \cite{Hong,10,20,25}) showing that a
single hidden layer perceptron with reasonably restricted set of weights
still retains the universal approximation property. But if weights are taken
from too \textquotedblleft narrow" sets, then the universal approximation
property is generally violated and there arisen the problem of
identification of compact sets $X\subset \mathbb{R}^{d}$ such that the
considered network approximates arbitrarily well any given continuous
function on $X$. In \cite{I1}, the first author considered the problem of
approximation by neural networks with weights varying on a finite set of
straight lines and thresholds from the whole set of real numbers. He
obtained a sufficient and also a necessary condition for well approximation
by such networks. The obtained conditions were proven to be not equivalent
and the problem of obtaining a necessary and at the same time sufficient
condition was remained unsolved.

The main purpose of this paper is to obtain necessary and at the same time
sufficient conditions for density of single hidden layer networks with
weights from a set of finitely many straight lines and thresholds from an
open interval. The obtained results (see Theorems 2.4 and 2.5) are based on
properties of so-called ridge functions and measures orthogonal to such
functions. These results determine the theoretical boundaries of efficacy of
the considered network model. That is, they characterize (in terms of
projective measures) compact sets $X\subset \mathbb{R}^{d}$, over which the
model preserves its general propensity to approximate arbitrarily well any
continuous multivariate function. We also prove that there exist smooth
activation functions for which the obtained density property holds even if
we keep the number of neurons fixed (see Theorem 2.6).

\begin{center}
\bigskip

\bigskip

{\large \textbf{2. Measure theoretic results}}
\end{center}

In this section, we give a sufficient and at the same time necessary
condition for approximation by single hidden layer perceptrons with weights
and thresholds from certain limited sets. More precisely, we allow weights
vary on finitely many directions and thresholds vary on an open interval.

Assume we are given $k$ directions $\mathbf{a}^{j}\in \mathbb{R}%
^{d}\backslash \{\mathbf{0\}}$, $j=1,...,k$. Let $L_{j}=\{t\mathbf{a}%
^{j}:t\in \mathbb{R}\},$ $j=1,...,k,$ and $W=\cup _{j=1}^{k}L_{j}.$ Let $%
\Theta $ be an open interval in $\mathbb{R}$ and $\sigma :\mathbb{%
R\rightarrow R}$ be a continuous activation function. Consider the following
set of single hidden layer perceptrons with weights from $W$ and thresholds
from $\Theta $:

\begin{equation*}
\mathcal{M}(\sigma ;W,\Theta )=span\{\sigma (\mathbf{w}\cdot \mathbf{x}%
-\theta ):~\mathbf{w}\in W,~\theta \in \Theta \}.\eqno(2.1)
\end{equation*}%
Note that elements of $\mathcal{M}(\sigma ;W,\Theta )$ are continuous
functions depending on the variable $\mathbf{x}\in \mathbb{R}^{d}$. We ask
and answer the following question. For which compact sets $X\subset \mathbb{R%
}^{d}$ the set of functions from $\mathcal{M}(\sigma ;W,\Theta )$, but
considered over $X$, is dense in $C(X)$? In the sequel, for a family of
continuous functions $K\subset C(\mathbb{R}^{d}),$ the notation $\overline{K}%
=C(X)$ (or the expression \textquotedblleft $K$ is dense in $C(X)$") will
mean that the restriction of this family to $X$ is dense in $C(X)$. Thus, we
want to characterize compact sets $X,$ for which $\overline{\mathcal{M}%
(\sigma ;W,\Theta )}=C(X).$ In the following, we give a measure theoretic
description of such sets.

We continue with the notion of image of a finite signed measure $\mu $ and a
measure space $(U,\mathcal{A},\mu ).$ Let $F$ be a mapping from the set $U$
to the set $T.$ Then a measure space $(T,\mathcal{B},\nu )$ is called an
image of the measure space $(U,\mathcal{A},\mu )$ if the measurable sets $%
B\in \mathcal{B}$ are the subsets of $T$ such that $F^{-1}(B)\in \mathcal{A}$
and

\begin{equation*}
\mu (F^{-1}(B))=\nu (B),\text{ for all }B\in \mathcal{B}\text{.}
\end{equation*}%
The measure $\nu $ is called an image of $\mu $ and denoted by $F\circ \mu $%
. Clearly,

\begin{equation*}
\left\Vert F\circ \mu \right\Vert \leq \left\Vert \mu \right\Vert ,
\end{equation*}%
since under mapping $F$ there is a possibility of mixing up the images of
those sets on which $\mu $ is positive with those where it is negative.
Besides, note that if a bounded function $g:T\rightarrow $ $\mathbb{R}$ is $%
F\circ \mu $-measurable, then the composite function $[g\circ
F]:U\rightarrow $ $\mathbb{R}$ is $\mu $-measurable and

\begin{equation*}
\int_{U}[g\circ F]d\mu =\int_{T}gd[F\circ \mu ].\eqno(2.2)
\end{equation*}

It is well known fact in Functional Analysis that for a compact Hausdorff
space $X$ and the space of continuous functions $C(X)$, a subspace $M\subset
C(X)$ is dense in $C(X)$ if and only if the only Borel measure orthogonal to
$M$ is the zero measure. We will use this fact in further analysis. In the
sequel, as a subspace $M$ we will take the set of linear combinations of
so-called ridge functions. A\textit{\ ridge function}, in its simplest
format, is a multivariate function of the form $g\left( \mathbf{a}\cdot
\mathbf{x}\right) $, where $g$ is a univariate function, $\mathbf{a}=\left(
a_{1},...,a_{d}\right) $ is a vector (direction) in $\mathbb{R}%
^{d}\backslash \{\mathbf{0}\}$, $\mathbf{x}=\left( x_{1},...,x_{d}\right) $
is the variable and $\mathbf{a}\cdot \mathbf{x}$ is the inner product. In
other words, a ridge function is a composition of a univariate function with
a linear functional over $\mathbb{R}^{d}.$ These functions arise naturally
in various fields. They arise in partial differential equations (where they
are called \textit{plane waves}, see, e.g., \cite{J}), in computerized
tomography (see, e.g., \cite{Kaz,Log,Sh}; the name ridge function was coined
by Logan and Shepp in \cite{Log}), in statistics (especially, in the theory
of projection pursuit and projection regression; see, e.g., \cite{C,F,H}),
in approximation theory (see, e.g., \cite{3,Is,Kroo,Lin,15,P,21}) and in
theory of neural networks (see, e.g., \cite{3,I1,14,P,20,S,X}). For a
systematic study of ridge functions, see the recently published monograph by
Pinkus \cite{Pinkus}. Note that the functions $\sigma (\mathbf{w}^{i}\mathbf{%
\cdot x}-\theta _{i})$ in (1.1) are ridge functions of the directions $%
\mathbf{w}^{i}$. Thus it is not surprising that many approximation theoretic
problems arisen in the theory of neural networks are usually reduced to the
corresponding problems in the field of ridge functions.

Assume we are given $k$ directions $\mathbf{a}^{1},...,\mathbf{a}^{k}.$
Consider the linear combinations of ridge functions

\begin{equation*}
\mathcal{R}\left( \mathbf{a}^{1},...,\mathbf{a}^{k}\right) =\left\{
\sum\limits_{i=1}^{k}g_{i}\left( \mathbf{a}^{i}\cdot \mathbf{x}\right)
:~g_{i}\in C(\mathbb{R)},~i=1,...,k\right\} .
\end{equation*}%
Note that we fix the directions $\mathbf{a}^{i}$ and vary only functions $%
g_{i}$. Thus $\mathcal{R}\left( \mathbf{a}^{1},...,\mathbf{a}^{k}\right) $
is a linear space. This space usually appear in mathematical problems of
computed tomography (see, e.g., \cite{Kaz,Log,Sh}).

Let $X$ be a compact subset of $\mathbb{R}^{d}.$ By $C^{\ast }(X)$ we denote
the class of regular measures defined on Borel subsets of $X.$ When a
measure $\mu \in C^{\ast }(X)$ is orthogonal to $\mathcal{R}\left( \mathbf{a}%
^{1},...,\mathbf{a}^{k}\right) $? That is, when

\begin{equation*}
\int_{X}fd\mu =0,
\end{equation*}%
for each function $f\in \mathcal{R}\left( \mathbf{a}^{1},...,\mathbf{a}%
^{k}\right) $? It follows immediately from (2.2) that $\mu $ is orthogonal
to $\mathcal{R}\left( \mathbf{a}^{1},...,\mathbf{a}^{k}\right) $ if and only
if the projective measures

\begin{equation*}
\mu _{\mathbf{a}^{i}}\overset{def}{=}\pi _{i}\circ \mu \equiv 0,\text{ }%
i=1,...,k,
\end{equation*}%
where $\pi _{i}$ are natural projections in the directions $\mathbf{a}^{i}$.
That is, $\pi _{i}(\mathbf{x})=\mathbf{a}^{i}\cdot \mathbf{x}$, $i=1,...,k.$
Thus we arrive at the following lemma.

\bigskip

\textbf{Lemma 2.1.} \textit{The subspace $\mathcal{R}\left( \mathbf{a}%
^{1},...,\mathbf{a}^{k}\right) $ is dense in $C(X)$ if and only if for any
regular Borel measure $\mu $ satisfying $\mu _{\mathbf{a}^{i}}\equiv 0,$ we
have $\mu \equiv 0$.}

\bigskip

Let us look at the hypothesis of Lemma 2.1 a little more closely. Consider a
finite set $\{\mathbf{x}^{1},\ldots ,\mathbf{x}^{n}\}\subset X$. For each $%
i=1,...,k,$ let the set $\{\mathbf{a}^{i}\cdot \mathbf{x}^{j}$, $j=1,...,n\}$
has $s_{i}$ different values, which we denote by $\gamma _{1}^{i},\gamma
_{2}^{i},...,\gamma _{s_{i}}^{i}.$ For each $i=1,...,k,$ and each $%
m=1,...,s_{i}$, consider the linear equation

\begin{equation*}
\sum_{j}\lambda _{j}=0,\eqno(2.3)
\end{equation*}%
where the sum is taken over all indices $j\in \{1,...,n\}$ such that $%
\mathbf{a}^{i}\cdot \mathbf{x}^{j}=\gamma _{m}^{i}.$ Combining all these
equations we have the system of $\sum_{i=1}^{n}s_{i}$ homogeneous linear
equations in $\lambda _{1},...,\lambda _{n}$. Note that the coefficients of
these equations are the integers $0$ and $1.$ If the system of equations
(2.3) has a solution $\{\lambda _{1},...,\lambda _{n}\}$ with $\lambda
_{j}\neq 0,$ $j=1,...,n$, then the finite set $\{\mathbf{x}^{1},\ldots ,%
\mathbf{x}^{n}\}$ is called a \textit{closed path} with respect to the
directions $\mathbf{a}^{1},...,\mathbf{a}^{k}$ (see \cite{8}). The existence
of closed paths means that the subspace $\mathcal{R}\left( \mathbf{a}%
^{1},...,\mathbf{a}^{k}\right) $ cannot be dense in $C(X).$ Indeed, with a
closed path $l=\{\mathbf{x}^{1},\ldots ,\mathbf{x}^{n}\}$ we can associate a
measure

\begin{equation*}
\mu _{l}=\sum_{j=1}^{n}\lambda _{j}\delta _{\mathbf{x}^{j}},\eqno(2.4)
\end{equation*}%
where $\delta _{\mathbf{x}^{j}}$ is a point mass at $\mathbf{x}^{j}$, $%
j=1,...,n$. Note that this measure is nonzero and orthogonal to the subspace
$\mathcal{R}\left( \mathbf{a}^{1},...,\mathbf{a}^{k}\right) $. Thus for
density $\overline{\mathcal{R}\left( \mathbf{a}^{1},...,\mathbf{a}%
^{k}\right) }=C(X),$ the set $X$ must not contain closed paths. For example,
let $k=3,$ $\mathbf{a}^{i},i=1,2,3,$ coincide with the coordinate directions
and $\varepsilon $ be a sufficiently small positive number. If $X$ contains
the set $l=\{(0,0,0),~(0,0,\varepsilon ),~(0,\varepsilon ,0),~(\varepsilon
,0,0),~(\varepsilon ,\varepsilon ,\varepsilon )\}$ (or its translation),
then for the nontrivial measure $\mu $ in (2.4) with the coefficients $%
\lambda _{j}$ equal to $-2,1,1,1,-1$ respectively, we have $\mu _{\mathbf{a}%
^{i}}=0,$ $i=1,2,3.$ That is, $\mu \in \mathcal{R}^{\perp }\left( \mathbf{a}%
^{1},\mathbf{a}^{2},\mathbf{a}^{3}\right) $ but $\mu \neq 0$; hence $%
\overline{\mathcal{R}\left( \mathbf{a}^{1},\mathbf{a}^{2},\mathbf{a}%
^{3}\right) }\neq C(X)$. The above analysis shows that by Lemma 2.1, density
may hold only for specific sets $X$ with no interior.

For the case $k=2,$ the condition \textquotedblleft for any regular Borel
measure $\mu $ satisfying $\mu _{\mathbf{a}^{i}}\equiv 0,$ we have $\mu
\equiv 0$" is geometrically well understood. In this case, we can paraphrase
the condition in terms of measures induced by lightning bolts. A \textit{%
lightning bolt} (or concisely, a \textquotedblleft bolt") with respect to
two directions $\mathbf{a}^{1}$ and $\mathbf{a}^{2}$ is a finite or infinite
ordered set of points $(\mathbf{p}^{1},\mathbf{p}^{2},...)$ in $\mathbb{R}%
^{d}$ with $\mathbf{p}^{i}\neq $ $\mathbf{\ p}^{i+1}$ and its units $\mathbf{%
p}^{i+1}-\mathbf{p}^{i}$ alternatively perpendicular to the directions $%
\mathbf{a}^{1}$ and $\mathbf{a}^{2}$ (see \cite{11,M}). A finite bolt $%
\left( \mathbf{p}^{1},...,\mathbf{p}^{n}\right) $ is said to be closed if $n$
is an even number and the set $\left( \mathbf{p}^{1},...,\mathbf{p}^{n},%
\mathbf{p}^{1}\right) $ also forms a bolt. Bolts are geometrically explicit
objects and can be efficiently used in questions of density. It is easy to
see that if $\overline{\mathcal{R}\left( \mathbf{a}^{1},\mathbf{a}%
^{2}\right) }=C(X),$ then $X$ does not contain closed bolts. Indeed, on a
closed bolt $\left( \mathbf{p}^{1},...,\mathbf{p}^{2n}\right) $ one can
construct the nontrivial measure

\begin{equation*}
\mu =\sum_{j=1}^{2n}(-1)^{j+1}\delta _{\mathbf{p}^{j}},
\end{equation*}%
for which $\mu _{\mathbf{a}^{1}}\equiv 0$ and $\mu _{\mathbf{a}^{2}}\equiv 0$%
. Nonexistence of closed bolts is a very simple and natural necessary
condition for the density $\overline{\mathcal{R}\left( \mathbf{a}^{1},%
\mathbf{a}^{2}\right) }=C(X).$ However, this condition is not sufficient.
There are highly nontrivial cases, when $X$ does not contain closed bolts
and at the same time $\overline{\mathcal{R}\left( \mathbf{a}^{1},\mathbf{a}%
^{2}\right) }\neq C(X).$ For example, assume $X$ has the following
properties:

(1) $X$ is the union of two disjoint closed sets $X_{1}$ and $X_{2}$ such
that $\mathbf{a}^{i}\cdot X_{1}=\mathbf{a}^{i}\cdot X_{2},$ $i=1,2.$

(2) $X$ does not contain closed bolts.

Note that there exist compact sets with these properties (see, e.g., \cite%
{11}). From the properties (1)-(2) it follows that $X$ contains an infinite
bolt, which we denote by $l=(\mathbf{p}^{1},\mathbf{p}^{2},...)$. With this
bolt we associate the sequence of measures

\begin{equation*}
\mu _{n}=\frac{1}{n}\sum_{j=1}^{n}(-1)^{j+1}\delta _{\mathbf{p}^{j}},\text{ }%
n=1,2,...\eqno(2.5)
\end{equation*}

Obviously, $\left\Vert \mu _{n}\right\Vert =1,$ for each $n.$ By the
Banach-Alaoglu theorem (see \cite[p. 66]{R}), the sequence $\{\mu
_{n}\}_{n=1}^{\infty }$ has a weak$^{\text{*}}$ limit point $\mu ^{\ast }.$
We claim that $\mu ^{\ast }\in \mathcal{R}^{\perp }\left( \mathbf{a}^{1},%
\mathbf{a}^{2}\right) $ and $\mu ^{\ast }\neq 0.$ Indeed from the definition
of bolts and (2.5) it follows that

\begin{equation*}
\left\vert \int_{X}(g_{1}+g_{2})d\mu _{n}\right\vert \leq \frac{2}{n}%
(\left\Vert g_{1}\right\Vert +\left\Vert g_{2}\right\Vert ),
\end{equation*}%
for any $g_{1}+g_{2}=g_{1}\left( \mathbf{a}^{1}\cdot \mathbf{x}\right)
+g_{2}\left( \mathbf{a}^{2}\cdot \mathbf{x}\right) $ in $\mathcal{R}\left(
\mathbf{a}^{1},\mathbf{a}^{2}\right) .$ Tending $n$ to infinity, we obtain
that

\begin{equation*}
\int_{X}(g_{1}+g_{2})d\mu ^{\ast }=0,
\end{equation*}%
and since $g_{1}+g_{2}$ is an arbitrary sum from $\mathcal{R}\left( \mathbf{a%
}^{1},\mathbf{a}^{2}\right) ,$ we conclude that $\mu ^{\ast }\in \mathcal{R}%
^{\perp }\left( \mathbf{a}^{1},\mathbf{a}^{2}\right) .$ The assertion that $%
\mu ^{\ast }\neq 0$ is obvious, since for a continuous function $%
f:X\rightarrow \mathbb{R}$ with the properties $f(x)=1$ for all $x\in X_{1}$
and $f(x)=-1$ for all $x\in X_{2}$, we have

\begin{equation*}
\left\vert \int_{X}fd\mu _{n}\right\vert =1
\end{equation*}%
and hence

\begin{equation*}
\left\vert \int_{X}fd\mu ^{\ast }\right\vert =1.
\end{equation*}

There are (worthy of consideration) cases when existence of infinite bolts
does not violate the density condition of Lemma 2.1. First note that the
relation $\mathbf{x}\thicksim \mathbf{y}$ when $\mathbf{x}$ and $\mathbf{y}$
belong to some bolt in a given compact set $X\subset \mathbb{R}^{d}$ defines
an equivalence relation. The equivalence classes are called \textit{orbits}.
If orbits of $X\subset \mathbb{R}^{d}$ are topologically closed sets, then$\
\overline{\mathcal{R}\left( \mathbf{a}^{1}{,}\mathbf{a}^{2}\right) }=C(X)$
is equivalent to nonexistence of closed bolts in $X$ (see \cite{M}). For
example, let $k=2$, $\mathbf{a}^{1}=(1,1)$, $\mathbf{a}^{2}=(1,-1)$ and $X\ $%
be the set
\begin{equation*}
\left\{ (0,0),(1,-1),(0,-2),(-1\frac{1}{2},-\frac{1}{2}),(0,1),(\frac{3}{4},%
\frac{1}{4}),(0,-\frac{1}{2}),(-\frac{3}{8},-\frac{1}{8}),(0,\frac{1}{4}),(%
\frac{3}{16},\frac{1}{16}),...\right\} .
\end{equation*}%
Note that $X$ is an infinite bolt forming a single orbit. Since this orbit
is a closed set, we obtain that $\overline{\mathcal{M}(\sigma ;W,\Theta )}%
=C(X)$.

The following proposition is a special case of the known general result of
Marshall and O'Farrell \cite{M} established for the sum of two algebras.

\bigskip

\textbf{Proposition 2.2.} \textit{Assume $X$ is a compact subset of $\mathbb{%
R}^{d}$. The space $\mathcal{R}\left( \mathbf{a}^{1}{,}\mathbf{a}^{2}\right)
$ is dense in $C(X)$ if and only if $X$ does not contain closed bolts and $%
\mu _{n}$ converges weak$^{\text{*}}$ to zero for each infinite bolt $(%
\mathbf{p}^{1},\mathbf{p}^{2},...)\subset X.$}

\bigskip

We have discussed the hypothesis of Lemma 2.1. That is, we analyzed the
situations when any regular Borel measure $\mu $ satisfying $\mu _{\mathbf{a}%
^{i}}\equiv 0$ is the trivial measure. Our analysis is summarized in the
following corollary.

\bigskip

\textbf{Corollary 2.3.} \textit{Assume we are given a compact set $%
X\subset \mathbb{R}^{d}$ and directions $\mathbf{a}^{1},...,\mathbf{a}%
^{k}\in \mathbb{R}^{d}\backslash \{\mathbf{0}\}.$}

\textit{1) Let $k>2$. If for any regular Borel measure $\mu $ in $C^{\ast
}(X)$ satisfying $\mu _{\mathbf{a}^{i}}\equiv 0,$ we have $\mu \equiv 0$,
then necessarily\textit{\ }the set $X$ contains no closed paths.}

\textit{2) Let $k=2.$ For any regular Borel measure $\mu $ in $C^{\ast }(X)$
satisfying $\mu _{\mathbf{a}^{i}}\equiv 0,$ we have $\mu \equiv 0$ if and
only if the set $X$ does not contain closed bolts and for any infinite bolt $%
(\mathbf{p}^{1},\mathbf{p}^{2},...)$ (if such exist), the sequence of
measures $\mu _{n}$ in (2.5) converges weak$^{\text{*}}$ to zero. If orbits
of $X$ are closed, then $\mu _{n}$ always converges weak$^{\text{*}}$ to
zero.}

\bigskip

\textbf{Remark.} Note that Corollary 2.3 is complete only in the case $k=2$.
When trying to generalize (unclosed) lightning bolts to the case $k>2$,
we face big combinatorial difficulties. We do not know a reasonable
description of a sequence of points $(\mathbf{p}^{1},\mathbf{p}^{2},...)$
and weighted point mass measures $\mu _{n}$ (similar to (2.5)) such that any
weak$^{\text{*}}$ limit point of $\mu _{n}$ is orthogonal to $\mathcal{R}%
\left( \mathbf{a}^{1},...,\mathbf{a}^{k}\right) $. For $k>2$, the
complete description of measures orthogonal to $\mathcal{R}\left( \mathbf{a}%
^{1},...,\mathbf{a}^{k}\right) $ seems to be beyond the scope of methods
discussed herein. Even in the simplest case when $k=d$ and $\mathbf{a}^{i}$
are the coordinate directions in $\mathbb{R}^{d}$, this problem is open (see,
e.g., \cite{11,M}).

\bigskip

Now we apply the above material to the problem of approximation by neural
networks $\mathcal{M}(\sigma ;W,\Theta )$ (see (2.1)). The following theorem
is valid.

\bigskip

\textbf{Theorem 2.4.} \textit{Assume $\sigma :\mathbb{R\rightarrow R}$ is a
continuous nonpolynomial activation function and $X$ is a compact subset of $%
\mathbb{R}^{d}$. Then the set $\mathcal{M}(\sigma ;W,\Theta )$ is dense in $%
C(X)$ if and only if for any measure $\mu \in C^{\ast }(X)$ satisfying $\mu
_{\mathbf{a}^{i}}\equiv 0,$ we have $\mu \equiv 0$.}

\begin{proof} \textbf{Sufficiency.} Let for any measure $\mu \in C^{\ast }(X)$
satisfying $\mu _{\mathbf{a}^{i}}\equiv 0,$ we have $\mu \equiv 0$. Then by
Lemma 2.1, the set $\mathcal{R}\left( \mathbf{a}^{1},...,\mathbf{a}%
^{k}\right) $ is dense in $C(X)$. This means that for any function $f\in
C(X) $ and any positive real number $\varepsilon $ there exist continuous
univariate functions $g_{i},$ $i=1,...,k,$ such that

\begin{equation*}
\left\vert f(\mathbf{x})-\sum_{i=1}^{k}g_{i}\left( \mathbf{a}^{i}{\cdot }%
\mathbf{x}\right) \right\vert <\frac{\varepsilon }{k+1}\eqno(2.6)
\end{equation*}%
for all $\mathbf{x}\in X$. Since $X$ is compact, the sets $Y_{i}=\{\mathbf{a}%
^{i}{\cdot }\mathbf{x:\ x}\in X\},\ i=1,2,...,k,$ are also compacts. Note
that if $\sigma $ is not a polynomial, then the set

\begin{equation*}
span\text{\ }\{\sigma (ty-\theta ):\ t\in \mathbb{R},~\theta \in \Theta \}
\end{equation*}%
is dense in $C(\mathbb{R)}$ in the topology of uniform convergence (see \cite%
{20}). This density result means that for the given $\varepsilon $ there
exist numbers $c_{ij},t_{ij}\in \mathbb{R},$ $\theta _{ij}\in \Theta $, $%
i=1,2,...,k$, $j=1,...,m_{i}$ such that%
\begin{equation*}
\left\vert g_{i}(y)-\sum_{j=1}^{m_{i}}c_{ij}\sigma (t_{ij}y-\theta
_{ij})\right\vert \,<\frac{\varepsilon }{k+1}\eqno(2.7)
\end{equation*}%
for all $y\in Y_{i},\ i=1,2,...,k.$ From (2.6) and (2.7) we obtain that

\begin{equation*}
\left\Vert f(\mathbf{x})-\sum_{i=1}^{k}\sum_{j=1}^{m_{i}}c_{ij}\sigma (t_{ij}%
\mathbf{a}^{i}{\cdot }\mathbf{x}-\theta _{ij})\right\Vert
_{C(X)}<\varepsilon .\eqno(2.8)
\end{equation*}%
Hence $\overline{\mathcal{M}(\sigma ;W,\Theta )}=C(X).$

\bigskip

\textbf{Necessity.} Let $X$ be a compact subset of $\mathbb{R}^{d}$ and the
set $\mathcal{M}(\sigma ;W,\Theta )$ be dense in $C(X).$ Then for an
arbitrary positive real number $\varepsilon $, inequality (2.8) holds with
some coefficients $c_{ij},t_{ij}$ and $\theta _{ij},\ i=1,2,\ j=1,...,m_{i}.$
Since for each $i=1,2,...,k$, the function $\sum_{j=1}^{m_{i}}c_{ij}\sigma
(t_{ij}\mathbf{a}^{i}{\cdot }\mathbf{x}-\theta _{ij})$ is of the form $g_{i}(%
\mathbf{a}^{i}{\cdot }\mathbf{x}),$ the subspace $\mathcal{R}\left( \mathbf{a%
}^{1},...,\mathbf{a}^{k}\right) $ is dense in $C(X)$. Then by Lemma 2.1, for
any measure $\mu \in C^{\ast }(X)$ satisfying $\mu _{\mathbf{a}^{i}}\equiv
0, $ we have $\mu \equiv 0$. This completes the proof of Theorem 2.4. \end{proof}

\bigskip

For the case $k=2,$ by applying Proposition 2.2 one can obtain the following
theorem.

\bigskip

\textbf{Theorem 2.5.} \textit{Assume $\sigma :\mathbb{R\rightarrow R}$ is a
continuous nonpolynomial activation function, $X$ is a compact subset of $%
\mathbb{R}^{d}$ and $k=2.$ Then the set $\mathcal{M}(\sigma ;W,\Theta )$ is
dense in $C(X$) if and only if $X$ does not contain closed bolts and the
measures $\mu _{n}$ in (2.5) converges weak$^{\text{*}}$ to zero for each
infinite bolt $(\mathbf{p}^{1},\mathbf{p}^{2},...)\subset X$.}

\bigskip

\textbf{Examples.}

(1) Let $X\subset \mathbb{R}^{d}$ be a compact set with an interior point.
Then in an open neighborhood (lying in $X$) of this point we can construct a
nontrivial measure of the form (2.4). Thus by Theorem 2.4, $\overline{%
\mathcal{M}(\sigma ;W,\Theta )}\neq C(X)$.

(2) Let we are given directions $\{\mathbf{a}^{j}\}_{j=1}^{k}$ and a compact
curve $\gamma $ in $\mathbb{R}^{d}$ such that for any $c\in \mathbb{R}$, $%
\gamma $ has at most one common point with each of the hyperplanes $\mathbf{a%
}^{j}\cdot \mathbf{x}=c$, $j=1,...,r.$ Then, clearly, for any measure $\mu
\in C^{\ast }(X)$, the condition $\mu _{\mathbf{a}^{i}}\equiv 0$ implies $%
\mu \equiv 0$. Thus, $\mathcal{M}(\sigma ;W,\Theta )$ is dense in $C(X)$.

(3) Let $k=2$ and $X$ be the union of two parallel line segments not
perpendicular to the given directions $\mathbf{a}^{1}$ and $\mathbf{a}^{2}$.
Then $X$ does not contain closed bolts and infinite bolts. Therefore, by
Theorem 2.5., $\overline{\mathcal{M}(\sigma ;W,\Theta )}=C(X)$.

\bigskip

The above theorems mean that to have a density result we have a wide choice;
we must chose the activation function $\sigma $ from the set of
\textquotedblleft nonpolynomials". Although we were able to make some
restrictions on weights and thresholds, the number of neurons in the single
hidden layer may grow depending on approximation accuracy. That is, if the
conditions of Theorem 2.4 are satisfied, then for given $f\in C(X)$ and $%
\varepsilon >0$ there exists a network $h\in \mathcal{M}(\sigma ;W,\Theta )$
such that
\begin{equation*}
\left\Vert f-h\right\Vert <\varepsilon .
\end{equation*}%
But this inequality does not mean that the number of neurons in $h$ is
uniformly bounded for all $f$ and $\varepsilon $. Note that in applications
it is necessary to define how many neurons one should take in hidden layers.
The more the number of neurons, the more the probability of the network to
give precise results. But unfortunately, practicality decreases with
increase of neurons in hidden layers. There are several results aimed at
this problem, which show that neural networks with two and more hidden
layers can approximate arbitrarily well any given continuous multivariate
function even if the number of neurons in hidden layers are uniformly
bounded (see, e.g., \cite{6,14,20}).

The following result allows us to make restrictions not only on the set of
weights but also on the number of neurons of approximating single hidden
layer networks. It turns out that among nonpolynomials there exists smooth
activation functions $\sigma $ for which density holds, even if we have a
limited number of neurons in the single hidden layer.

By $\mathcal{M}_{r}(\sigma ;W;\mathbb{R})$ we will denote the set of
networks in $\mathcal{M}(\sigma ;W;\mathbb{R})$ with exactly $r$ neurons in
a hidden layer. That is,
\begin{equation*}
\mathcal{M}_{r}(\sigma ;W;\mathbb{R})=\left\{ \sum_{i=1}^{r}c_{i}\sigma (%
\mathbf{w}^{i}\mathbf{\cdot x}-\theta _{i}):\text{ }c_{i}\in \mathbb{R},%
\text{ }\mathbf{w}^{i}\in W,\text{ }\theta _{i}\in \mathbb{R},\text{ }%
i=1,...,r\right\} .
\end{equation*}

\bigskip

\textbf{Theorem 2.6.} \textit{Let $X$ be a compact subset of $\mathbb{R}^{d}$%
. Let, besides, $\mathbf{a}^{j}\in \mathbb{R}^{d}\backslash \{\mathbf{0\}}$,
$j=1,...,k$, and $W=\cup _{j=1}^{k}\{t\mathbf{a}^{j}:t\in \mathbb{R}\}.$
Then there exists an infinitely differentiable activation function $\sigma :$
$\mathbb{R}\rightarrow \mathbb{R}$, for which the following statement is
valid: the set $\mathcal{M}_{k}(\sigma ;W,\mathbb{R})$ is dense in $C(X)$ if
and only if for any measure $\mu \in C^{\ast }(X)$ satisfying $\mu _{\mathbf{%
a}^{i}}\equiv 0,$ we have $\mu \equiv 0$.}

\begin{proof} \textbf{Sufficiency.} Let for any measure $\mu \in C^{\ast }(X)$
satisfying $\mu _{\mathbf{a}^{i}}\equiv 0,$ we have $\mu \equiv 0$. Then by
Lemma 2.1, the set $\mathcal{R}\left( \mathbf{a}^{1},...,\mathbf{a}%
^{k}\right) $ is dense in $C(X)$. This means that for any function $f\in
C(X) $ and any positive real number $\varepsilon $ there exist continuous
univariate functions $g_{i},$ $i=1,...,k,$ for which the inequality (2.6) is
satisfied. Since $X$ is compact, the sets $Y_{i}=\{\mathbf{a}^{i}\cdot
\mathbf{x:\ x}\in X\},\ i=1,...,k,$ are also compact. Assume all the sets $%
Y_{i}$ are contained in a line segment $[-l,l].$

Let $\alpha $ be any number. Divide the interval $[\alpha ,+\infty )$ into
the segments $[\alpha ,2\alpha ],$ $[2\alpha ,3\alpha ],...$. We construct $%
\sigma ,$ obeying the hypothesis of Theorem 2.5, in two stages. Let $%
\{p_{m}(t)\}_{m=1}^{\infty }$ be the sequence of polynomials with rational
coefficients defined on $[-l,l].$ First, we define $\sigma $ on the closed
intervals $[(2m-1)\alpha ,2m\alpha ],$ $m=1,2,...$ as the function

\begin{equation*}
\sigma (t)=p_{m}(\frac{2l}{\alpha }t-4ml+l),\text{ }t\in \lbrack
(2m-1)\alpha ,2m\alpha ],
\end{equation*}%
or equivalently,

\begin{equation*}
\sigma (\frac{\alpha }{2l}t+2m\alpha -\frac{\alpha }{2})=p_{m}(t),\text{ }%
t\in \lbrack -l,l].\eqno(2.9)
\end{equation*}

At the second stage we extend $\sigma $ to the intervals $(2m\alpha
,(2m+1)\alpha ),$ $m=1,2,...,$ and $(-\infty ,\alpha )$, maintaining the $%
C^{\infty }$ property.

For any univariate function $g\in C(\mathbb{R})$ and any $\varepsilon >0$
there exists a polynomial $p(t)$ with rational coefficients such that

\begin{equation*}
\left\vert g(t)-p(t)\right\vert <\frac{\varepsilon }{k},
\end{equation*}%
for all $t\in \lbrack -l,l].$ This together with (2.9) mean that

\begin{equation*}
\left\vert g(t)-\sigma (\frac{\alpha }{2l}t-r)\right\vert <\frac{\varepsilon
}{k},
\end{equation*}%
for some $r\in \mathbb{R}$ and all $t\in \lbrack -l,l].$ Thus for the
functions $g_{i}$ in (2.6) we can write that

\begin{equation*}
\left\vert g_{i}(\mathbf{a}^{i}{\cdot }\mathbf{x})-\sigma (\lambda \mathbf{a}%
^{i}{\cdot }\mathbf{x}-r_{i})\right\vert <\frac{\varepsilon }{k},\eqno(2.10)
\end{equation*}%
where $\lambda =\frac{\alpha }{2l}.$ From (2.6) and (2.10) we obtain that
\begin{equation*}
\left\vert f(\mathbf{x})-\sum_{i=1}^{k}\sigma (\lambda \mathbf{a}^{i}{\cdot }%
\mathbf{x}-r_{i})\right\vert <\varepsilon \eqno(2.11)
\end{equation*}%
The inequality (2.11) can be written in the form

\begin{equation*}
\left\vert f(\mathbf{x})-\sum_{i=1}^{k}c_{i}\sigma (\mathbf{w}^{i}{\cdot }%
\mathbf{x}-\theta _{i})\right\vert <\varepsilon ,\eqno(2.12)
\end{equation*}%
where $\mathbf{w}^{i}\in W,$ $c_{i}=1$ and $\theta _{i}$ are some real
numbers. This means that the set $\mathcal{M}_{k}(\sigma ;W,\mathbb{R})$ is
dense in $C(X)$.

\bigskip

\textbf{Necessity.} Let $X$ be a compact subset of $\mathbb{R}^{d}$ and
assume the set $\mathcal{M}_{k}(\sigma ;W,\mathbb{R})$ is dense in $C(X).$
Then for any positive real number $\varepsilon $, the inequality
(2.12) holds with some coefficients $c_{i},\theta _{i}\in \mathbb{R}$ and $%
\mathbf{w}^{i}\in W,$ for $i=1,...,k.$ Since the function $%
\sum_{i=1}^{k}c_{i}\sigma (\mathbf{w}^{i}{\cdot }\mathbf{x}-\theta _{i})$ is
a function of the form $\sum_{i=1}^{k}g_{i}(\mathbf{a}^{i}{\cdot }\mathbf{x}%
),$ the subspace $\mathcal{R}\left( \mathbf{a}^{1},...,\mathbf{a}^{k}\right)
$ is dense in $C(X)$. Then by Lemma 2.1, for any measure $\mu \in C^{\ast
}(X)$ satisfying $\mu _{\mathbf{a}^{i}}\equiv 0,$ we have $\mu \equiv 0$.
\end{proof}

\end{document}